\documentclass[lettersize,journal]{IEEEtran}
\usepackage{amsmath,amsfonts}
\usepackage{algorithmic}
\usepackage{algorithm}
\usepackage{array}
\usepackage[caption=false,font=normalsize,labelfont=sf,textfont=sf]{subfig}
\usepackage{textcomp}
\usepackage{stfloats}
\usepackage{url}
\usepackage{verbatim}
\usepackage{graphicx}
\usepackage{cite}
\usepackage{float}
\usepackage{booktabs}
\usepackage{multirow}
\usepackage{colortab}
\usepackage{subfig}
\begin{document}

\title{UVL2: A Unified Framework for Video Tampering Localization}

\author{
    \IEEEauthorblockN{Pengfei Pei$^{1,2}$, Yun Cao$^{*1,2}$, Jinchuan Li$^{1,2}$, Zeyu Zhang$^{1,2}$, Yuqi Pang$^{1,2}$}\\
    \IEEEauthorblockA{$^1$ Institute of Information Engineering, Chinese Academy of Sciences, Beijing, China}\\
    \IEEEauthorblockA{$^2$ School of Cyber Security, University of Chinese Academy of Sciences, Beijing, China}
    \IEEEauthorblockA{\{peipengfei,caoyun,lijinchuan,zhangzeyu,pangyuqi\}@iie.ac.cn}
}



\maketitle

\begin{abstract}
With the advancement of deep learning-driven video editing technology, security risks have emerged. Malicious video tampering can lead to public misunderstanding, property losses, and legal disputes. Currently, detection methods are mostly limited to specific datasets, with limited detection performance for unknown forgeries, and lack of robustness for processed data. This paper proposes an effective video tampering localization network that significantly improves the detection performance of video inpainting and splicing by extracting more generalized features of forgery traces. Considering the inherent differences between tampered videos and original videos, such as edge artifacts, pixel distribution, texture features, and compress information, we have specifically designed four modules to independently extract these features. Furthermore, to seamlessly integrate these features, we employ a two-stage approach utilizing both a Convolutional Neural Network and a Vision Transformer, enabling us to learn these features in a local-to-global manner. Experimental results demonstrate that the method significantly outperforms the existing state-of-the-art methods and exhibits robustness.
\end{abstract}


\section{Introduction}

With the development of deep learning, it has become increasingly easy to produce realistic synthetic videos, making it difficult for people to visually detect the traces of video tampering \cite{VIFST,MVSS-Net,FAST}. In this paper, we investigate video tampering localization, including the detection of video inpainting and video splicing. Video inpainting typically eliminates objects that originally existed in the video, leading people to believe that the objects never existed \cite{VIFST,VIDNet,MVSS-Net,FAST}. On the other hand, video splicing adds an additional object to the original video, leading people to believe that the object originally existed \cite{MVSS-Net,pei2023uvl,VideoSplicingDetection}. When these technologies are maliciously used, they may lead to misunderstandings and cause serious consequences, especially in the political and military arenas. 

In the pursuit of detecting tampered regions within videos, researchers have introduced a range of efficient detection algorithms. These algorithms aim to enhance detection accuracy by capturing nuanced feature discrepancies between tampered and authentic regions \cite{VideoSplicing_2024,MVSS-Net,VIFST,hpf,VIDNet,Sub-RegionHashingRetrieval}. These methodologies encompass approaches such as error-level analysis \cite{VIDNet}, tracking spatial and temporal trajectories \cite{DVIL2022}, and the design of specialized feature enhancement networks \cite{LIN2023,LJC_VI}. For the detection of object splicing, researchers have further leveraged techniques such as JPEG compression artifacts, optical flow inconsistencies, high-pass filtering, boundary artifacts, and pixel noise estimation \cite{ComNet,flow-edge,hpf,MVSS-Net,GSR-Net}.

However, these methods are often tailored to specific forgery types or datasets, resulting in inefficiencies and limited generalization when faced with the diverse tampering methods and emerging synthesis techniques in the real world. To address these challenges, we propose a generalized video object tampering localization algorithm. Given that video splicing and inpainting essentially involve the synthesis of two non-homologous video segments, either genuine or AI-generated, the tampered regions inherently exhibit inconsistencies with the original regions. This inconsistency provides a theoretical basis for our algorithmic approach.

In practice, videos undergo various post-processing techniques when distributed online, such as resolution adjustment, compression, blurring, image enhancement, and cropping\cite{MVSS-Net,VIFST,FAST,ViTHash}. Existing methods that rely solely on a single feature tend to suffer a significant reduction in performance when detecting unknown types of tampered data, especially when such data has undergone these post-processing steps. As a result, they often fail to generalize effectively and robustly.

Therefore, to improve the generalization performance and robustness of our algorithm for unknown data, it is crucial to conduct thorough research on the forgery traces left by splicing or inpainting tampering. This involves devising a generalizable feature that is resilient to various video tampering methods and post-processing techniques. The proposed algorithm in this paper aims to achieve this by leveraging the inherent inconsistencies between tampered and original regions in videos.

To learn features that are independent of video synthesis methods, we investigate common inherent traces of various types of tampered videos. Since tampered videos are essentially composed of two non-homogeneous videos, the original and tampered regions come from different sources, leading to various inconsistencies. These inconsistencies include: (1) Texture features: It is difficult for two videos from different sources to be completely consistent in natural conditions. Texture information can help analyze the natural environment at the time of filming in different regions. (2) Edge features: Splicing videos or tampering areas often leave uneven boundaries at the edges. Analyzing edge features can effectively improve the detection accuracy. (3) Pixel and noise distribution: Due to the physical imaging principles of camera capture, the noise distribution of each device or AI-generated image is different. Analyzing these pixel or physical noise differences can improve pixel-level detection accuracy. (4) Other tampering traces: Some other traces are not easily detected through various spatial changes, such as compression information. These not easily detected tampering features can be effectively solved by frequency domain characteristics. To extract these features, we designed four corresponding modules.

After extracting the tampering video traces in different aspects such as texture, edge, noise, and frequency domain, in order to comprehensively learn and utilize these features, we need to introduce an efficient network structure. Convolutional Neural Network (CNN) has excellent performance in extracting local features of images, which can better capture local information and details in images\cite{HRFormer, VIFST}. In contrast, Vision Transformer (ViT) is more suitable for sequence data learning and can better adapt to the temporal structure of videos. ViT can extract the global contextual correlation of features. Based on these characteristics, we adopt a two-stage feature extraction method that combines the advantages of CNN and ViT. First, we use a CNN-based model to extract local features of the video, making the feature scale smaller. Then, we use a ViT-based structure to capture the contextual relationship between these local features. Finally, we output the localization results of the tampered area in the video.

\begin{figure}[t]
    \begin{center}
        \includegraphics[width=1.\linewidth]{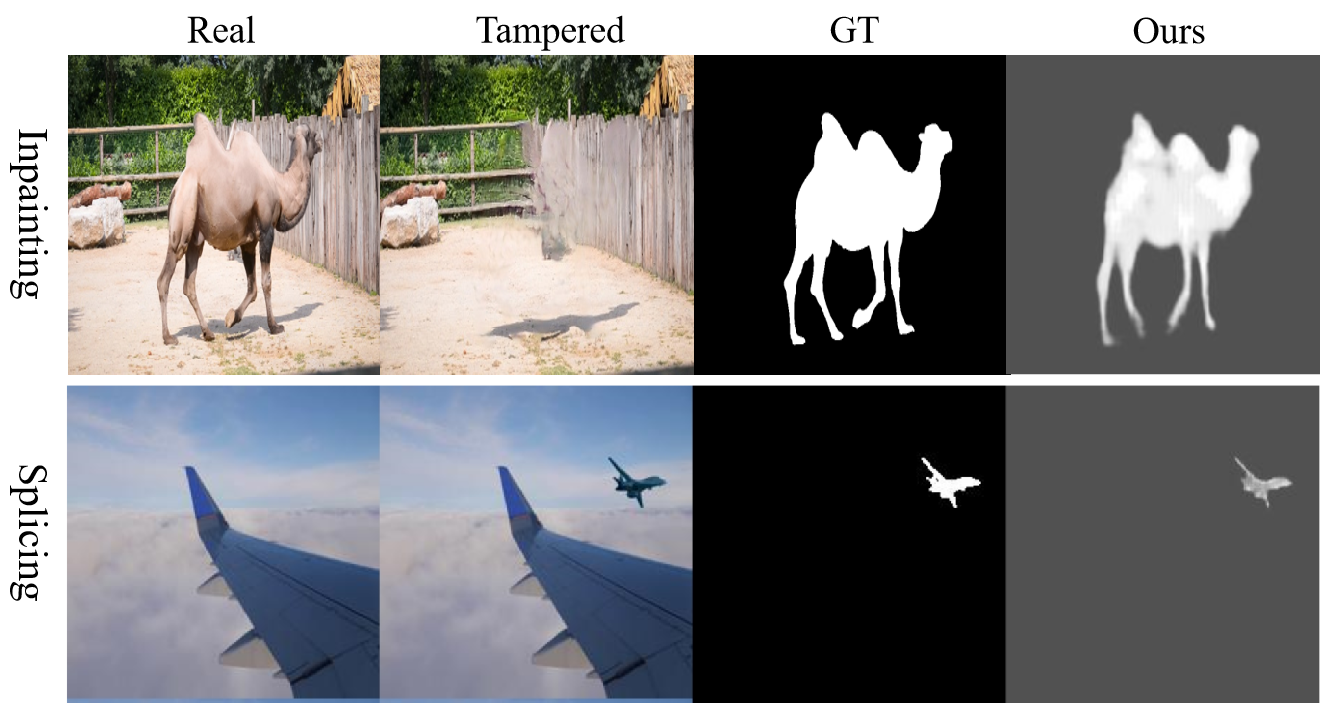}
    \end{center}
    \caption{
        The localization of tampered areas in video inpainting and video splicing.
    }
    \label{fig:uvl-localization}
\end{figure}
Figure \ref{fig:uvl-localization} demonstrates the video tampering localization results achieved by the proposed method, validating its excellent performance in overall detection and intricate detail handling. In summary, the key contributions of this paper are as follows:

\begin{itemize}
\item We introduce a general and efficient video tampering localization network that significantly enhances video detection efficiency and effectively applies to the detection of video inpainting and splicing.
\item We design a multi-view feature extraction module, enabling the proposed method to capture more generalized features, thus enhancing the network's ability to generalize for unknown non-homologous data.
\item Extensive experiments across diverse datasets for video inpainting and video splicing tampering localization, have demonstrated that the proposed method exhibits superior generalization and robustness, outperforming current comparative methods.
\end{itemize}

\section{Related Works}
\begin{figure*}[t]
    \centering
    \includegraphics[width=\textwidth]{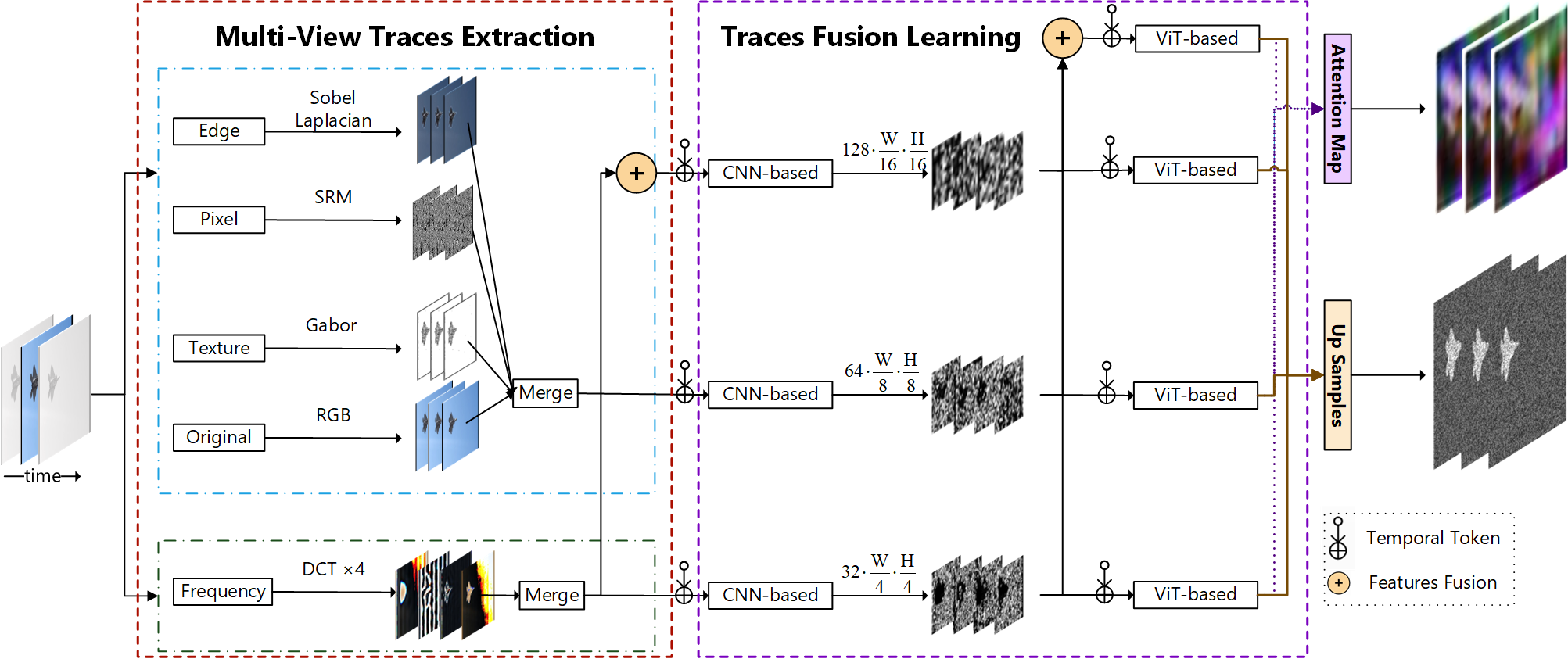}
    \caption{Overview of the framework. We first extract inconsistencies between the original and tampered regions of the video from four aspects: texture, edges, pixels, and frequency domain. Then, we adopt a two-stage learning structure based on CNN and ViT to achieve correlation learning from local to global, which is used to fuse these inconsistent features. Finally, the output is a video consisting of pixel-level localization results of the tampered region. It's worth noting that within each stage, the Features Fusion module combines features and generates an additional feature branch for input to the next stage.}
    \label{fig:arch-vis}
\end{figure*}

\subsection{Video Inpainting Localization}

Recent research has witnessed remarkable progress in video inpainting localization techniques. Zhang et al.\cite{LIN2023} advanced the field by enhancing artifact modules through the concatenation of a feature extractor and forgery output module, enabling a deeper learning of inpainting features and resulting in more precise localization of inpainting regions. Zhou et al.\cite{VIDNet} pioneered a dual-stream encoder-decoder architecture, incorporating attention modules and LSTM, to detect restored video regions. Their approach leverages error level analysis and temporal structure information to accurately identify inpainted areas.

Wei et al.\cite{DVIL2022} took a different approach, focusing on analyzing spatial-temporal traces left by inpainting operations. They enhanced these traces using intra-frame and inter-frame residuals, guided by optical flow, and then utilized a dual-stream encoder and bidirectional LSTM decoder to predict inpainted regions.

Lou et al. \cite{lou2024videoinpaintinglocalizationcontrastive} introduced a 3D Uniformer encoder to learn video noise residuals, effectively capturing spatio-temporal forensic features. They further employed supervised contrastive learning to highlight local inconsistencies in repaired videos, improving localization performance. In a follow-up work \cite{lou2024trustedvideoinpaintinglocalization}, Lou et al. explored a multi-scale noise extraction module based on 3D high-pass layers to create noise modalities. A cross-modal attention fusion module was utilized to explore the correlation between these modalities, and an attentive noise decoding module selectively enhanced spatial details, further enhancing the network's localization capabilities.

Given the diversity of object inpainting methods, enhancing the generalization capability of detecting various object-restored videos in a computer vision context is imperative. This capability is crucial for addressing the variations introduced by different inpainting techniques and ensuring accurate detection across various scenarios, thereby providing a more reliable safeguard for the authenticity and integrity of video content.

\subsection{Video Splicing Localization}

The field of video splicing detection, particularly at the object level, has witnessed significant advancements in recent years. Techniques such as PQMECNet\cite{JPEG} have emerged, leveraging local estimation of JPEG primary quantization matrices to distinguish spliced regions originating from different sources. MVSS-Net\cite{MVSS-Net}, on the other hand, learns semantically irrelevant yet generalizable features by analyzing noise distributions and boundary artifacts surrounding tampered regions. ComNet\cite{ComNet} customizes approximate JPEG compression operations to enhance detection performance against JPEG compression artifacts.

DCU-Net\cite{DCU-Net} further advances this field by effectively extracting tampered regions through dual-channel encoding fusion of deep features, the utilization of dilated convolutions, and the integration of decoders. Additionally, TransU2-Net\cite{TransU2-Net} proposes a novel hybrid transformer architecture specifically designed to improve object-level forgery detection in images. MSA-Net\cite{MSA-Net-ImageSplicing}, in particular, introduces a multi-scale attention network that integrates a multi-scale self-attention mechanism, enabling it to capture global dependencies and fine-grained details, resulting in precise localization of various types and sizes of spliced forged objects.

Yadav et al.\cite{yadav2024visuallyattentivesplicelocalization} also contribute to this field with a visually attentive image splicing localization network that comprises a Multi-Domain Feature Extractor (VA-MDFE) and a Multi-Receptive Field Upsampler (VA-MRFU). The VA-MDFE extracts features from RGB, edge, and depth modalities, while the VA-MRFU upsamples these features using multi-receptive field convolutions. Several researchers have successfully applied semantic segmentation techniques to the task of splicing localization, achieving notable results\cite{Fairness-Attention24,RethinkingAttention2024,MVSS-Net}.

However, the challenge in splicing localization lies in the common video manipulations often found on social media platforms, such as compression, cropping, detail enhancement, and blurring. These manipulations can disrupt the tampering traces of the original forged videos, significantly reducing the robustness of existing detection methods. Therefore, further research is needed to develop splicing detection techniques that are more resilient to such manipulations.

\section{Method}

As shown in Figure \ref{fig:arch-vis}, our method mainly consists of two core components: multi-view tampering trace extraction and tampering trace fusion learning. We have designed a series of modules to extract features from different perspectives, including texture, edge, pixel and frequency domain. These features are used to analyze the differences between the spliced or inpainted objects and the original video regions. In the tampering trace fusion learning component, we adopt CNN-based and ViT-based modules to process these features, thus achieving relevant learning from local to global. The CNN module focuses on extracting local features, especially good at dealing with edges, textures and other details. In contrast, the ViT module excels at extracting global features, which helps to understand the semantic information of the entire video. Through the joint learning of these two components, we are able to effectively solve the task of video tampering localization, improving detection accuracy and robustness. Finally, we generate a grayscale video of pixel-level localization results of tampering areas, where white represents tampering areas and gray represents original areas.

\subsection{Texture Feature Extraction Module}
When two video segments are spliced together, they often exhibit distinct textural characteristics due to their potentially differing capture locations or physical environments. Detecting these textural features in spliced videos is paramount for analyzing video authenticity. By comparing and analyzing the textural differences between video segments, we can assist in determining whether a video has undergone splicing or tampering. To extract textural features, this section utilizes CNN convolution kernels to implement Gabor filters. Gabor filters, through multi-scale and multi-directional analysis, can efficiently capture various textural information in images. They exhibit excellent sensitivity to textural features of different frequencies, directions, and polarities, making them suitable for a wide range of textural analysis tasks. The mathematical expressions of Gabor filters are given as follows:

\begin{align*}
G(x, y) &= \exp \left(-\frac{x'^2 + \gamma^2 y'^2}{2\sigma^2} \right) \cos \left(2\pi \frac{x'}{\lambda} + \phi \right) \\
G(x, y) &= \exp \left(-\frac{x'^2 + \gamma^2 y'^2}{2\sigma^2} \right) \sin \left(2\pi \frac{x'}{\lambda} + \phi \right)
\end{align*}

Here, $x$ and $y$ represent the spatial coordinates of the image, while $x'$ and $y'$ represent the coordinates after rotation and scaling transformations. $\sigma$ controls the bandwidth of the filter, $\lambda$ determines the central frequency of the filter, $\gamma$ is the decay factor, and $\phi$ represents the phase offset. By applying Gabor filters to input video frames, we can obtain textural responses at different scales and directions. These textural responses can be utilized to construct representations of textural features, which aid in detecting inconsistencies between objects and backgrounds at splicing junctions.

\subsection{Edge Feature Extraction Module}
\begin{figure*}[t]
    \centering
    \subfloat[Sobel]
    {
        \label{fig:sobel}
        \includegraphics[width=.45\linewidth]{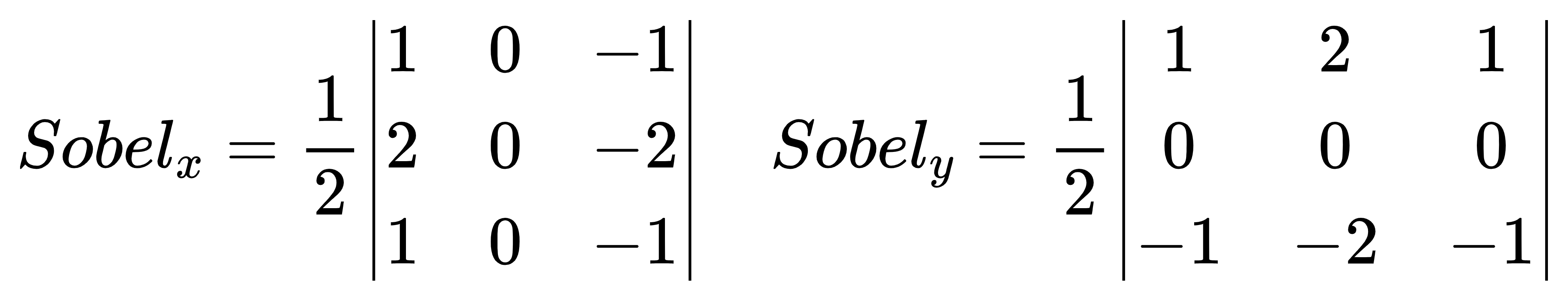}
   }
   \subfloat[Laplacian]
    {
        \label{fig:laplace}
        \includegraphics[width=.45\linewidth]{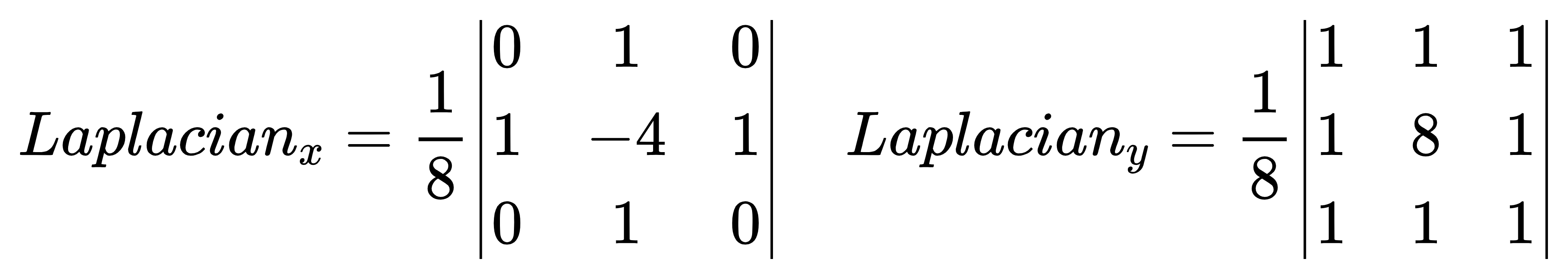}
   }\\
    \subfloat[SRM]
    {
        \label{fig:srm}
        \includegraphics[width=1.\linewidth]{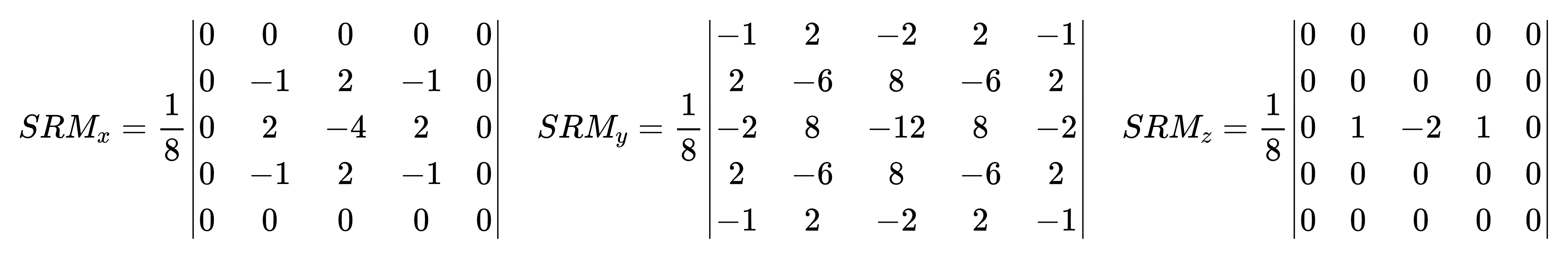}
   }
    \caption{
        The Sobel operator, Laplacian operator and SRM operator are used in the spatial domain branch.
        }
    \label{fig:ops}
\end{figure*}

In order to gain profound insight and comprehension of edge detection characteristics, we have deliberately employed two classical edge detection operators: the Sobel operator and the Laplacian operator. Sobel and Laplacian operators are commonly used for first-order and second-order edge detection, respectively. Sobel filters seek brightness changes in the image by calculating the first-order discrete derivative values of the grayscale function, particularly at edges. This helps highlight areas of brightness and lighting variations. Laplacian filters more precisely locate brightness changes and edges by calculating the second-order gradient of the image, while also enhancing details, including textures and small brightness variations.

As demonstrated in Figure \ref{fig:sobel} and \ref{fig:laplace}, we utilize the \textit{x} and \textit{y} components of these operators as convolutional kernels within a Convolutional Neural Network (CNN) during implementation. Specifically, these kernels are independently applied to the red (R), green (G), and blue (B) channels of the original video frames to extract edge features. To facilitate the training of the neural network, we further normalize these features to a range of 0-1. The Sobel operator achieves edge detection by computing the image intensity gradient for each pixel, while the Laplacian operator, based on the divergence of the scalar function gradient, exhibits rotational invariance, thus enabling more robust edge detection. The unique advantages of these two operators, when used in tandem, allow for the extraction of more robust and accurate edge features.

\subsection{Pixel Feature Extraction Module}

Due to imaging principles, pixel and noise distribution from different regions can vary. Therefore, analyzing these distributions is crucial. Noise info in videos is stored in pixels, and the Spatial Rich Model (SRM) operator can analyze pixel features. This algorithm relies on pixel statistical features like inter-pixel correlation, skewness, and kurtosis, which reflect the relationships between pixels. Manipulated regions using deep learning often disrupt natural pixel correlations. The SRM operator uses neighboring pixels' statistical features to represent local image structure.

By analyzing these statistical features, the SRM operator can better represent the destruction of multiple correlations between neighboring pixels in manipulated regions of the image, thereby improving the accuracy and generalization of image tampering detection. We apply the SRM operator to RGB frames, treating pixels in the red, green, and blue channels as three separate images. Each channel will yield a set of SRM features.

\subsection{Frequency Feature Extraction Module}
The extraction of frequency domain information is crucial for video quality analysis. The Discrete Cosine Transform (DCT) is a common method to convert video signals from the spatial domain to the frequency domain. It decomposes videos into different frequency components, and the amplitude and phase information of these components can be used to assess video quality. High amplitudes suggest strong signals, while low ones may indicate signal loss or noise. By analyzing this info, we can evaluate metrics like clarity and distortion levels, enhancing splicing tampering detection accuracy and robustness. This ensures video content's credibility and integrity. 

In our implemented frequency filter model, we utilize four distinct block sizes to process video information across various frequency ranges. These filters process: (1) low-frequency features, encompassing overall structure and larger features. (2) mid-frequency features, between low and high. (3) high-frequency features, capturing details and texture. (4) the entire spectrum, covering low, mid, and high frequencies, for transformations on the entire image. Finally, the outputs from these filters are concatenated to form the final output.


\subsection{Tampering Traces Fusion Learning}
The objective of multi-view fusion learning of tampering traces is to generate pixel-level prediction results that align with ground truth. To achieve this, we leverage the strengths of CNN and ViT models. We use the ResNet\cite{MVSS-Net} module for local feature extraction and the InterlacedFormer\cite{HRFormer} module for global feature extraction. By combining CNN and ViT, we gain a deeper understanding of image details and content. CNN excels at extracting local features, focusing on edges and textures, aiding in more precise tampering region localization. Meanwhile, ViT's strength lies in global feature extraction, enhancing overall video semantic understanding. By integrating global features, the model better understands the relationship between tampered objects and their environments, further improving localization accuracy. Ultimately, we produce pixel-level localization results for tampered regions in videos.

\section{Experiments}
\begin{table*}
\tabcolsep=12pt
	\centering
	\footnotesize
 \caption{Comparison experiments on the DAVIS-VI dataset with recent approaches that test on one subset and train on the remaining subsets (*).}
\label{tab:video-inpainting}
\resizebox{\linewidth}{!}{
\begin{tabular}{c|ccc|ccc|ccc} 
\hline \hline \noalign{\smallskip}
Methods      & \begin{tabular}[c]{@{}c@{}}VI*\\mIoU/F1\end{tabular} & \begin{tabular}[c]{@{}c@{}}OP*\\mIoU/F1\end{tabular} & \begin{tabular}[c]{@{}c@{}}CP\\mIoU/F1\end{tabular} & \begin{tabular}[c]{@{}c@{}}VI\\mIoU/F1\end{tabular} & \begin{tabular}[c]{@{}c@{}}OP*\\mIoU/F1\end{tabular} & \begin{tabular}[c]{@{}c@{}}CP*\\mIoU/F1\end{tabular} & \begin{tabular}[c]{@{}c@{}}VI*\\mIoU/F1\end{tabular} & \begin{tabular}[c]{@{}c@{}}OP\\mIoU/F1\end{tabular} & \begin{tabular}[c]{@{}c@{}}CP*\\mIoU/F1\end{tabular}  \\ 
\hline
NOI          & 0.08/0.14                                            & 0.09/0.14                                            & 0.07/ 0.13                                          & 0.08/0.14                                           & 0.09/0.14                                            & 0.07/0.13                                            & 0.08/0.14                                            & 0.09/0.14                                           & 0.07/0.13                                             \\
CFA          & 0.10/0.14                                            & 0.08/0.14                                            & 0.08/0.12                                           & 0.10/0.14                                           & 0.08/0.14                                            & 0.08/0.12                                            & 0.10/0.14                                            & 0.08/0.14                                           & 0.08/0.12                                             \\
COSNet       & 0.40/0.48                                            & 0.31/0.38                                            & 0.36/0.45                                           & 0.28/0.37                                           & 0.27/0.35                                            & 0.38/0.46                                            & 0.46/0.55                                            & 0.14/0.26                                           & 0.44/0.53                                             \\
HPF~         & 0.46/0.57                                            & 0.49/0.62                                            & 0.46/0.58                                           & 0.34/0.44                                           & 0.41/0.51                                            & 0.68/0.77                                            & 0.55/0.67                                            & 0.19/0.29                                           & 0.69/0.80                                             \\
HPF+LSTM     & 0.50/0.61                                            & 0.39/0.51                                            & 0.52/0.63                                           & 0.26/0.36                                           & 0.38/0.44                                            & 0.68/0.78                                            & 0.53/0.64                                            & 0.20/0.30                                           & 0.70/0.81                                             \\
GSR-Net~     & 0.57/0.68                                            & 0.50/0.63                                            & 0.51/0.63                                           & 0.30/0.43                                           & 0.74/0.80                                            & 0.80/0.85                                            & 0.59/0.70                                            & 0.22/0.33                                           & 0.70/0.77                                             \\
GSR-Net+LSTM & 0.55/0.67                                            & 0.51/0.64                                            & 0.53/0.64                                           & 0.33/0.45                                           & 0.60/0.72                                            & 0.74/0.83                                            & 0.58/0.70                                            & 0.21/0.32                                           & 0.71/0.81                                             \\
VIDNet~      & 0.55/0.67                                            & 0.46/0.58                                            & 0.49/0.63                                           & 0.31/0.42                                           & 0.71/0.77                                            & 0.78/0.86                                            & 0.58/0.69                                            & 0.20/0.31                                           & 0.70/0.82                                             \\
VIDNet-BN    & 0.62/0.73                                            & 0.75/0.83                                            & 0.67/0.78                                           & 0.30/0.42                                           & 0.80/0.86                                            & \textbf{0.84}/\textbf{0.92}                          & 0.58/0.70                                            & 0.23/0.32                                           & 0.75/0.85                                             \\
VIDNet-IN    & 0.59/0.70                                            & 0.59/0.71                                            & 0.57/0.69                                           & 0.39/0.49                                           & 0.74/0.82                                            & 0.81/0.87                                            & 0.59/0.71                                            & 0.25/0.34                                           & 0.76/0.85                                             \\
FAST         & 0.61/0.73                                            & 0.65/0.78                                            & 0.63/0.76                                           & 0.32/0.49                                           & 0.78/0.87                                            & 0.82/0.90                                            & 0.57/ 0.68                                           & 0.22/0.34                                           & 0.76/0.83                                             \\
TruVIL       & 0.61/0.72                                            & \textbf{0.82/0.89}                                   & 0.70/0.81                                        & 0.42/0.54                                           & 0.84/0.91                                            & 0.82/0.89                                            & 0.63/0.74                                            & 0.53/0.67                                           & 0.81/0.88                                             \\
ViLocal      & 0.66/0.77                                            & \textbf{0.82/0.89}                                   & 0.76/0.85                                           & 0.61/0.73                                           & \textbf{0.85/0.91}                                   & 0.83/0.90                                            & 0.69/0.79                                            & 0.63/0.75                                           & 0.82/0.89                                             \\
VIFST        & 0.73/0.84                                            & 0.81/0.89                                            & 0.72/0.82                                           & 0.75/0.85                                           & 0.72/0.83                                            & 0.82/0.89                                            & \textbf{\textbf{0.85/0.91}}                          & 0.79/0.87                                           & 0.84/0.90                                             \\ 

Ours         & \textbf{0.78/0.87}                                   & 0.78/0.87                                            & \textbf{0.76/0.85}                                  & \textbf{0.76/0.85}                                  & 0.82/0.89                                            & 0.83/0.89                                            & \textbf{0.85/0.91}                                   & \textbf{0.80/0.88}                                  & \textbf{0.85/0.91}                                    \\
\hline \hline
\end{tabular}
}
\end{table*}
\subsection{Experimental Setup}

\subsubsection{Dataset}
Yu et al. \cite{FAST} and VIFST \cite{VIFST} introduced the DAVIS-VI dataset, a video inpainting dataset based on the DAVIS dataset. To generate tampered videos, they utilized six video inpainting methods: OPN \cite{opn}, CPNET \cite{LeeOWK19}, DVI \cite{KimWLK19a}, FGVC \cite{flow-edge}, DFGVI \cite{XuLZL19}, and STTN \cite{ZengFC20}. The DAVIS-VI dataset comprises 50 original videos and 300 tampered videos, totaling 33,550 frames. The Video Splicing (VS) dataset\cite{ViTHash} was designed specifically for video splicing detection and consists of a training set with 795 forged videos and a test set with 30 carefully crafted forged videos and 30 real videos.

\subsubsection{Evaluation Metrics and Baselines}
To evaluate pixel-level manipulation localization, we employ two commonly used metrics: mean Intersection over Union (mIoU), and F1-score (F1). For video inpainting, the methods we compare include: NOI \cite{VIDNet}, CFA \cite{VIDNet}, CosNet \cite{FAST}, HPF \cite{hpf}, GSR-Net \cite{GSR-Net}, VIDNet \cite{VIDNet}, FAST \cite{FAST}, TruVIL \cite{lou2024trustedvideoinpaintinglocalization}, ViLocal \cite{lou2024videoinpaintinglocalizationcontrastive} and VIFST \cite{VIFST}. The results of all these comparative methods are referenced from VIFST \cite{VIFST}. For video splicing, the methods we compare are: PoolFormer \cite{MetaFormer}, MetaFormer \cite{MetaFormer}, HRFormer \cite{HRFormer}, UVL-Net \cite{pei2023uvl}, UVL-Net-S \cite{pei2023uvl}, and UVL-Net-F \cite{pei2023uvl}. We have retrained these methods on the VS dataset and validated their performance.

\subsection{Results of Video Inpainting Localization}

Table \ref{tab:video-inpainting} shows the pixel-level detection performance of various models. Compared to the baseline method, our method achieves superior performance in most cases, but it is only slightly inferior to VIDNet-BN and VIFST on one sub-dataset. At the same time, our method performs well in detecting all unknown forged data, which is mainly due to the ability of multi-view features to enhance the generalized feature extraction ability of our method. The experimental results demonstrate that our method is not only efficient, but also has good generalization ability.

\subsection{Results of Video Splicing Localization}
\begin{table*}
\tabcolsep=12pt
\centering
\footnotesize
\caption{Robustness experiment on the VS dataset .}
\label{tab:compare-vs}
\begin{tabular}{c|ccccccc} 
\hline \hline \noalign{\smallskip}
Methods & \begin{tabular}[c]{@{}c@{}}None\\mIoU/F1\end{tabular} & \begin{tabular}[c]{@{}c@{}}Compression\\mIoU/F1\end{tabular} & \begin{tabular}[c]{@{}c@{}}Detail\\mIoU/F1\end{tabular} & \begin{tabular}[c]{@{}c@{}}Gaussian\\mIoU/F1\end{tabular} & \begin{tabular}[c]{@{}c@{}}Blur\\mIoU/F1\end{tabular} & \begin{tabular}[c]{@{}c@{}}Median\\mIoU/F1\end{tabular} & \begin{tabular}[c]{@{}c@{}}Filp\\mIoU/F1\end{tabular}  \\ 
\hline
PoolFormer                   & 0.789/0.873                                           & 0.765/0.852                                                  & 0.786/0.871                                             & 0.736/0.824                                               & 0.760/0.847                                           & 0.787/0.872                                             & 0.777/0.864                                            \\
MetaFormer                   & 0.726/0.822                                           & 0.685/0.779                                                  & 0.727/0.823                                             & 0.645/0.736                                               & 0.667/0.761                                           & 0.720/0.815                                             & 0.711/0.809                                            \\
HRFormer                     & 0.788/0.868                                           & 0.769/0.851                                                  & 0.786/0.867                                             & 0.748/0.833                                               & 0.765/0.848                                           & 0.787/0.869                                             & 0.775/0.860                                            \\
UVL-Net                      & 0.783/0.867                                           & 0.757/0.843                                                  & 0.779/0.863                                             & 0.750/0.831                                               & 0.757/0.839                                           & 0.780/0.863                                             & 0.781/0.866                                            \\
UVL-Net-S                    & 0.797/0.877                                           & 0.770/0.854                                                  & 0.792/0.873                                             & 0.744/0.823                                               & 0.765/0.848                                           & 0.796/0.876                                             & 0.793/0.875                                            \\
UVL-Net-F                    & 0.796/0.877                                           & 0.779/0.863                                                  & 0.794/0.875                                             & 0.764/0.848                                               & \textbf{0.783/0.865}                                  & 0.798/0.878                                             & 0.786/0.868                                            \\ 
Ours                         & \textbf{0.804/0.883}                                  & \textbf{0.786/0.865}                                         & \textbf{0.798/0.878}                                    & \textbf{0.772/0.849}                                      & 0.781/0.860                                           & \textbf{0.802/0.881}                                    & \textbf{0.795/0.874}                                   \\
\hline \hline
\end{tabular}

\end{table*}
Table \ref{tab:compare-vs} presents comparative experiments with the latest methods and robustness testing results. Although related methods perform well, our method demonstrates superior performance in fine-grained pixel-level detection. This advantage may stem from our targeted analysis of video splicing forgery. Additionally, compared to other semantic segmentation methods such as HRFormer, PoolFormer, and MetaFormer, known video splicing detection methods exhibit significant advantages. This indicates that the analysis of forgery traces in these methods is effective. Meanwhile, the detection results for various processed videos, such as those that have undergone compression, blurring, cropping, etc., show that our method exhibits smaller variations in outcome ($<$ 0.03). Experimental results indicate that multi-view forgery trace features contribute to improving the performance and robustness of video splicing detection.

\subsection{Ablation Study}
\label{sec:ablation}
\begin{figure*}
\centering
    \subfloat[Different features]
    {
        \includegraphics[width=.5\linewidth]{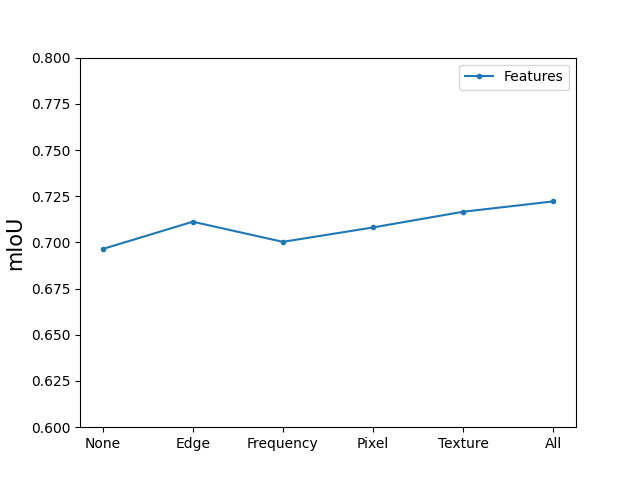}
        \label{fig:ab-bottles}  
    }
    \subfloat[Different components]
    {
        \includegraphics[width=.5\linewidth]{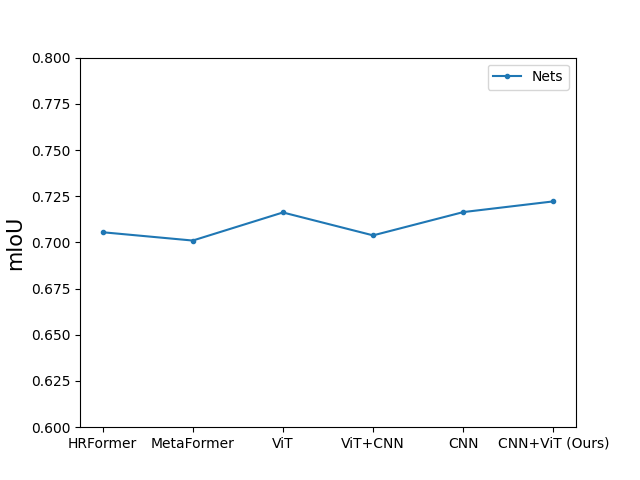}
        \label{fig:ab-comps}  
    }
    \caption{
    Ablation study on the proposed network on the DAVIS-VI dataset to assess the impact of different features and components. 
    }
    \label{fig:ab-vis}
\end{figure*}

\subsubsection{Impact of Various Features}
The results in Figure \ref{fig:ab-bottles} demonstrate the detection outcomes when applying different features. The ``All" result, which incorporates the four features mentioned in this paper -- edge, pixel, texture, and frequency domain features -- achieved the best result. This indicates that combining multiple features can effectively enhance detection accuracy. 

The ``Edge" module, which utilizes Sobel and Laplacian operators to effectively extract first- and second-order edge features, demonstrates superior performance compared to the ``None" baseline. This indicates that edge features are effective in detecting forged regions.

The ``Pixel" module, when compared to the ``None" baseline, also exhibits superior performance. This suggests that inconsistencies in pixel and noise distribution can aid in detecting differences between original and forged regions. 

Similarly, when compared to the ``None" baseline, the ``Texture" module successfully extracts the different texture between two regions, leading to improved performance.

Lastly, the ``Frequency" module, which applies DCT to extract different compression artifacts between two regions, also demonstrates superior performance compared to the ``None" baseline. This indicates that the ``Frequency" module is effective.

\subsubsection{The influence of different components}
The figure referred to as Figure \ref{fig:ab-comps} validates the effectiveness of our designed network structure for the multi-view forgery trace fusion learning component. Specifically, our CNN+ViT(Ours) outperforms other methods that utilize CNNs and ViTs. When compared to traditional advanced network architectures like MetaFormer and HRFormer, these methods incorporate CNNs into ViTs as a holistic approach to reduce ViT computational overhead for building large-scale models. However, our approach differs in that we utilize CNNs and ViTs in two separate stages. The results demonstrate that this approach is more suitable for video tamper detection and leads to the best performance, albeit with increased memory consumption.

When we only use one of CNN or ViT to learn multi-view features, the performance is lower than CNN+ViT. This is because relying solely on CNN can lead to a lack of global correlation information, which affects pixel-level accuracy. Conversely, using ViT alone can limit sensitivity to local features, resulting in poor detection performance. Experimental results confirm that combining CNN and ViT in this way is effective.

We also evaluated our method by reversing the order of ViT and CNN. Surprisingly, ViT+CNN performed worse than when using either alone. This suggests that first analyzing local tampering traces using CNN and then leveraging ViT to learn global correlations is more beneficial for detecting tampering regions. This aligns with our expectations and common sense.

\subsection{Visualization Results Analysis and Discussion}
\begin{figure*}[ht]
\centering
    \subfloat[Video inpainting]
    {
        \includegraphics[width=.5\linewidth]{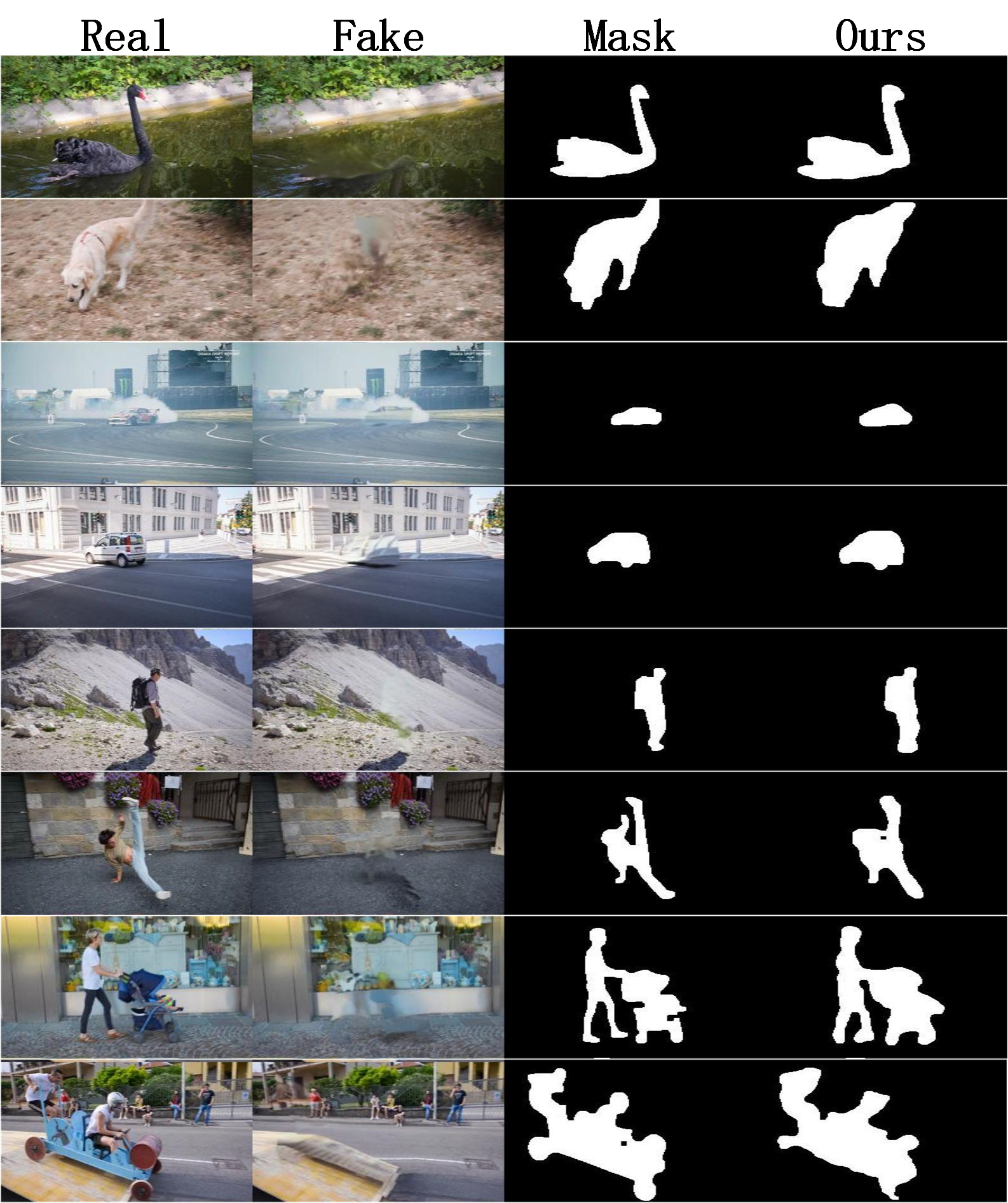}
        \label{fig:vi-results}  
    }
    \subfloat[Video splicing]
    {
        \includegraphics[width=.5\linewidth]{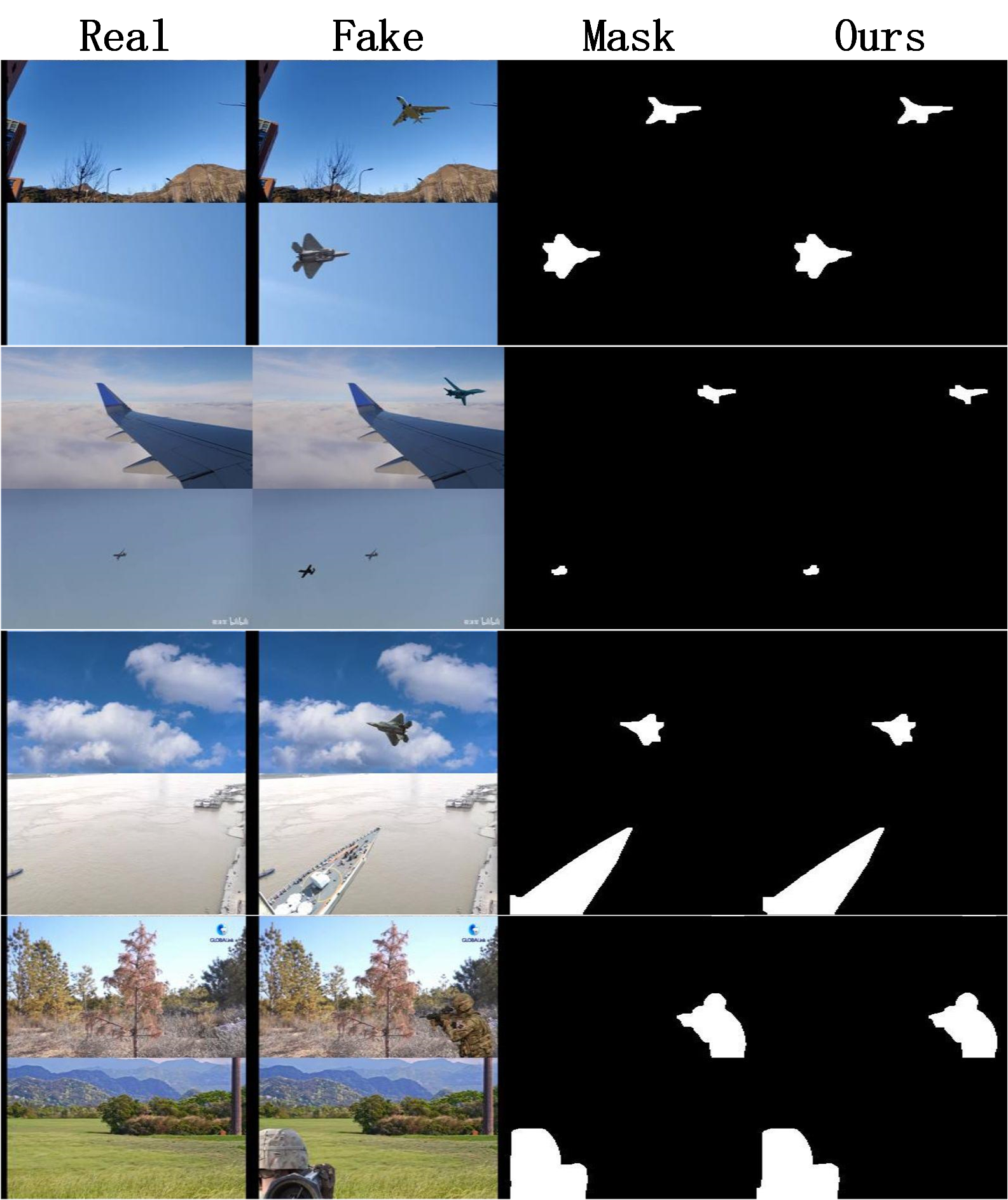}
        \label{fig:vs-results}  
    }
    \caption{
        The detection results on the DAVIS-VI and VS dataset are presented. In the results, ``Real" represents the original video, ``Fake" indicates the tampered video, ``Mask" indicates the ground truth, and ``Ours" represents the detection results of our method.
    }
    \label{fig:vis-results}
\end{figure*}

Figure \ref{fig:vis-results} illustrates the localization results of video inpainting and video splicing alterations. Figure \ref{fig:vi-results} present the results of video inpainting detection, including animals, cars, humans, and combined objects. Our detection for single-object alterations is excellent, even for complex objects. Figure \ref{fig:vs-results} demonstrates video object splicing detection with military objects such as aircrafts, tanks, warships, and soldiers. We can easily detect these alterations through analysis of various forgery traces. This indicates that our method is effective for video alteration detection. Through the analysis of multi-view alteration signature characteristics, we have achieved good results in fine-grained detection such as edges, small components, and combined objects. However, there is still room for improvement in the localization of complex alteration objects. In the future, we will strive to enhance our ability to detect alterations of multiple combined objects.

\section{Conclusion}

In this paper, we have presented an effective video tampering localization strategy that significantly improves the detection performance of video inpainting and splicing. With the advancement of deep learning-driven video editing technology, the emergence of security risks, such as malicious video tampering, has become a pressing concern. We address this issue by extracting more generalized features of forgery traces, considering the inherent differences between tampered videos and original videos. To effectively fuse these features, we adopt a CNN+ViT two-stage method to learn them in a local-to-global manner. Experimental results demonstrate that our method significantly outperforms the existing state-of-the-art methods and exhibits robustness. Our future work will focus on extending this localization strategy to other types of video tampering and exploring more advanced feature extraction methods to further improve detection performance.

\section*{Acknowledgments}
This should be a simple paragraph before the References to thank those individuals and institutions who have supported your work on this article.

\bibliographystyle{IEEEtran}
\bibliography{main}

\vfill

\end{document}